\documentclass[11pt]{article}

\usepackage[final]{acl}

\usepackage{times}
\usepackage{latexsym}
\usepackage{float}

\usepackage{algorithm}
\usepackage{algpseudocode}
\usepackage{amsmath}
\usepackage{amssymb}

\usepackage[T1]{fontenc}

\usepackage[utf8]{inputenc}

\usepackage{microtype}

\usepackage{inconsolata}

\usepackage{graphicx}
\usepackage{inconsolata}
\usepackage{booktabs}
\usepackage{soul}
\usepackage{color}
\usepackage{balance}

\usepackage{multirow}
\usepackage{multicol}
\usepackage{enumitem,kantlipsum}
\usepackage[normalem]{ulem}

\usepackage{color,colortbl}
\usepackage{todonotes} 
\usepackage{soul}
\usepackage{threeparttable, makecell}
\usepackage{makecell}
\usepackage{xcolor}

\usepackage{comment}
\usepackage{array}  
\usepackage{lscape}  
\usepackage{listings} %
\usepackage{multirow}
\usepackage{graphicx}
\usepackage[normalem]{ulem}
\useunder{\uline}{\ul}{}

\usepackage{array}  
\usepackage{multirow}  
\usepackage{amsmath}
\usepackage{subcaption}
\usepackage{fancybox}
\usepackage{tikz}

\definecolor{codegray}{gray}{0.9}

\lstdefinestyle{pythonstyle}{
    backgroundcolor=\color{codegray},   
    language=Python,
    basicstyle=\ttfamily\footnotesize,
    keywordstyle=\color{blue},
    stringstyle=\color{red},    
    breaklines=true,
    frame=single,
    keepspaces=true,
    showstringspaces=false,
}

\lstdefinestyle{aclprompt}{
  basicstyle=\ttfamily\small,
  columns=fullflexible,
  breaklines=true,
  breakatwhitespace=true,
  showstringspaces=false,
  keepspaces=true,
  frame=single,
  framerule=0.4pt,
  rulecolor=\color{black!25}
}

\lstdefinestyle{promptstyle}{
  basicstyle=\ttfamily\footnotesize,
  breaklines=true,
  breakatwhitespace=false,
  columns=fullflexible,
  keepspaces=true,
  showstringspaces=false,
  frame=single,
  framerule=0.3pt,
  rulecolor=\color{black!25},
  xleftmargin=0.5em,
  xrightmargin=0.5em,
  aboveskip=0.6em,
  belowskip=0.6em
}

\makeatletter
\renewcommand{\footnoterule}{%
  \kern-3pt
  \hrule width 0.4\columnwidth
  \kern 2.6pt
}
\makeatother
\usepackage{xcolor}

\usepackage{graphicx}
\definecolor{lightblue}{rgb}{.50,.95,1}
\definecolor{tri}{rgb}{.25,.88,.82}
\definecolor{lilac}{rgb}{0.85,0.64,0.85}

\usepackage{ulem}
\usepackage{hyperref}
\usepackage{verbatim}
\usepackage{setspace}

\usepackage{colortbl}
\usepackage{booktabs}

\usepackage[edges]{forest}
\usepackage{forest}

\usetikzlibrary{arrows.meta}
\tikzset{>=Latex}
\usepackage{cuted}

\usepackage{pifont}
\newcommand{\cmark}{\ding{51}} %

\usepackage{fontawesome5}
\usepackage[table]{xcolor}
\usepackage{pifont}

\definecolor{AvailBg}{RGB}{232,245,233} %
\definecolor{ProgBg}{RGB}{255,243,224}  %

\definecolor{Prog}{RGB}{227,114,34}     %

\usepackage{booktabs}
\usepackage{pifont}
\usepackage{fontawesome5}
\usepackage[table]{xcolor}
\usepackage{array}
\usepackage{makecell}

\definecolor{AvailBg}{RGB}{236,248,238} %
\definecolor{ProgBg}{RGB}{255,247,235}  %
\definecolor{Prog}{RGB}{227,114,34}

\newcommand{\avail}{\cellcolor{AvailBg}\cmark}
\newcommand{\inprog}{\cellcolor{ProgBg}\textcolor{Prog}{\faHourglassHalf}}
\newcommand{\availlegend}{\colorbox{AvailBg}{\cmark}}
\newcommand{\inproglegend}{\colorbox{ProgBg}{\textcolor{Prog}{\faHourglassHalf}}}

\title{\textbf{CritiSense}: Critical Digital Literacy and
Resilience Against Misinformation}

\author{
Firoj Alam$^{1}$,
Fatema Ahmad$^{1}$,
Ali Ezzat Shahroor$^{1}$,
Mohamed Bayan Kmainasi$^{1}$\\
{\bf
Elisa Sartori$^{2}$,
Giovanni Da San Martino$^{2}$,
Abul Hasnat$^{3}$,
Raian Ali$^{4}$ 
}\\
$^{1}$Qatar Computing Research Institute, Qatar,
$^{2}$University of Padova, Italy \\
$^{3}$APAVI.AI, France, 
$^{4}$Hamad Bin Khalifa University, Qatar \\
\texttt{fialam@hbku.edu.qa}\\
\url{https://critisense-web.digitqr.net/}
}

\begin{document}
\maketitle

\begin{strip}
\vspace{-6mm} %
\centering
\includegraphics[width=\textwidth,height=0.32\textheight,keepaspectratio]{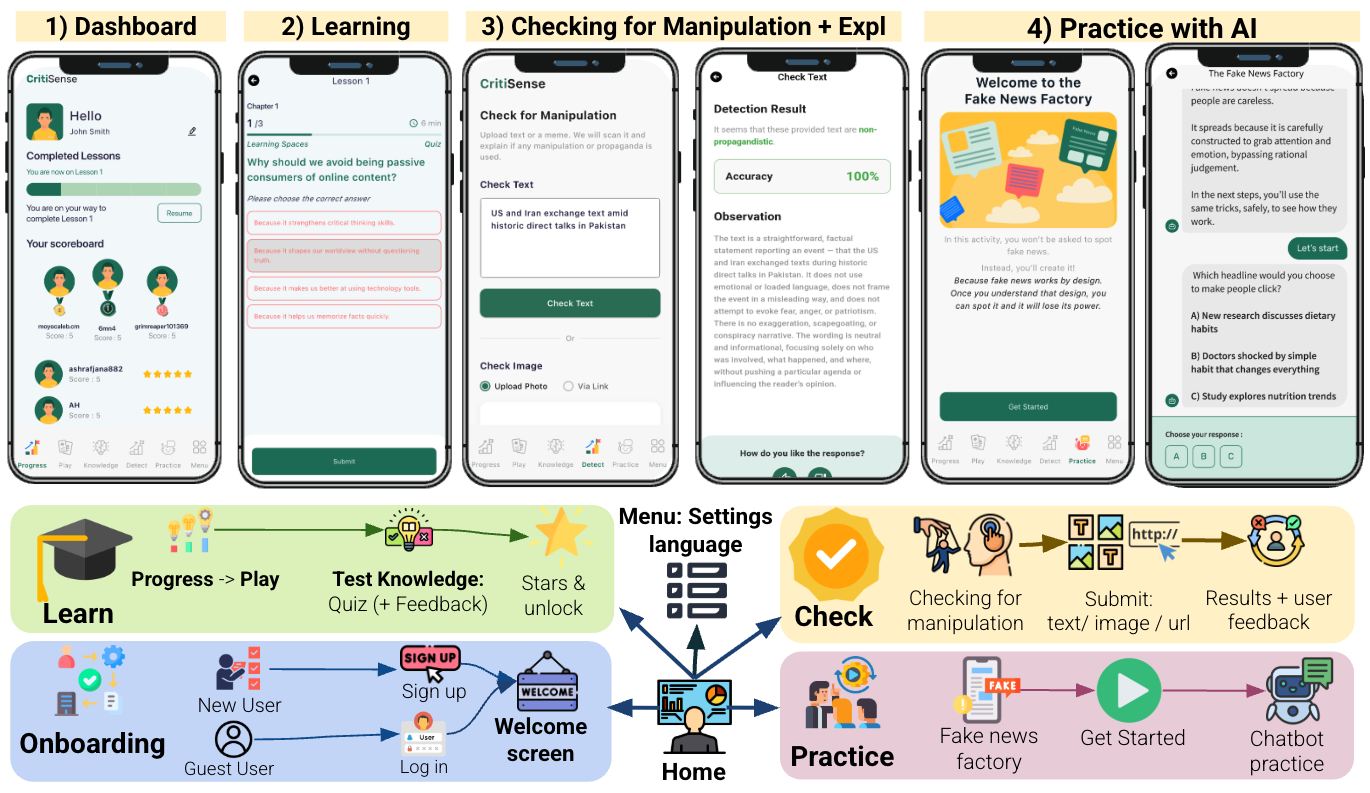}
\captionsetup{type=figure,skip=20pt}
\caption{Overview of \textsc{CritiSense} app and its functionalities.}
\label{fig:system_overview}
\end{strip}

\begin{abstract}
Misinformation on social media undermines informed decision-making and public trust. Prebunking offers a proactive complement by helping users recognize manipulation tactics \emph{before} they encounter them in the wild. We present \textsc{CritiSense}, a mobile media-literacy app that builds these skills through short, interactive challenges with instant feedback. It is the \textit{first} multilingual (supporting nine languages) and modular platform, designed for rapid updates across topics and domains. We report a usability study with 93 users: 83.9\% expressed overall satisfaction and 90.1\% rated the app as easy to use. Qualitative feedback indicates that \textsc{CritiSense} helps improve digital literacy skills. Overall, it provides a multilingual prebunking platform and a testbed for measuring the impact of microlearning on misinformation resilience.
Over 6 months, we have reached 500+ active users. It is freely available %
on the Apple \href{https://apps.apple.com/us/app/critisense/id6749675792}{App Store} and Google \href{https://play.google.com/store/apps/details?id=com.critisense&hl=en}{Play Store}.

\end{abstract}

\section{Introduction}
\label{sec:introduction}

The rapid diffusion of online misinformation threatens public health, democratic governance, and social cohesion. Since false claims often spread faster than corrections, post-hoc fact-checking can arrive too late and may suffer from fatigue and limited behavioral impact. Reflecting the scale of this challenge, the World Economic Forum’s Global Risks Report 2026 ranks mis- and disinformation among the world’s top near-term risks (second on the two-year outlook)~\cite{wef_grr2026}. To address these challenges, social media platforms rely on automatic detection, fact-checking pipelines, and warning interfaces to curb misleading content. While these measures provide essential first-line protection, they are largely reactive and claim-specific, can degrade under temporal and cross-lingual/domain shift, and do not by themselves build durable user competence \cite{berger2025_debunking,stepanova-ross-2023-temporal}. 

In practice, detection-centric mitigation is most effective when complemented with user-facing training, for three technical reasons: \textbf{\textit{(i)} Temporal drift:} misinformation is non-stationary, narratives and multimodal presentation styles evolve quickly, producing temporal distribution shift that can challenge deployed classifiers; temporally out-of-domain evaluation shows that performance can drop even for strong models \citep{stepanova-ross-2023-temporal}. \textbf{\textit{(ii)} Data and coverage:} robust detection benefits from large, high-quality annotations, yet such resources are uneven across languages and regions, and cross-lingual transfer remains less reliable for lower-resource settings \citep{ozcelik-etal-2023-cross}. \textbf{\textit{(iii)} Interface design:} detection outputs must be translated into interventions (e.g., labels, banners, downranking), where careful UI choices are crucial to maximize critical evaluation and avoid unintended effects, such as ``implied truth'' for untagged items \citep{pennycook-etal-2020-implied-truth}.

\begin{figure}[]
    \centering    
    \includegraphics[width=0.98\columnwidth]{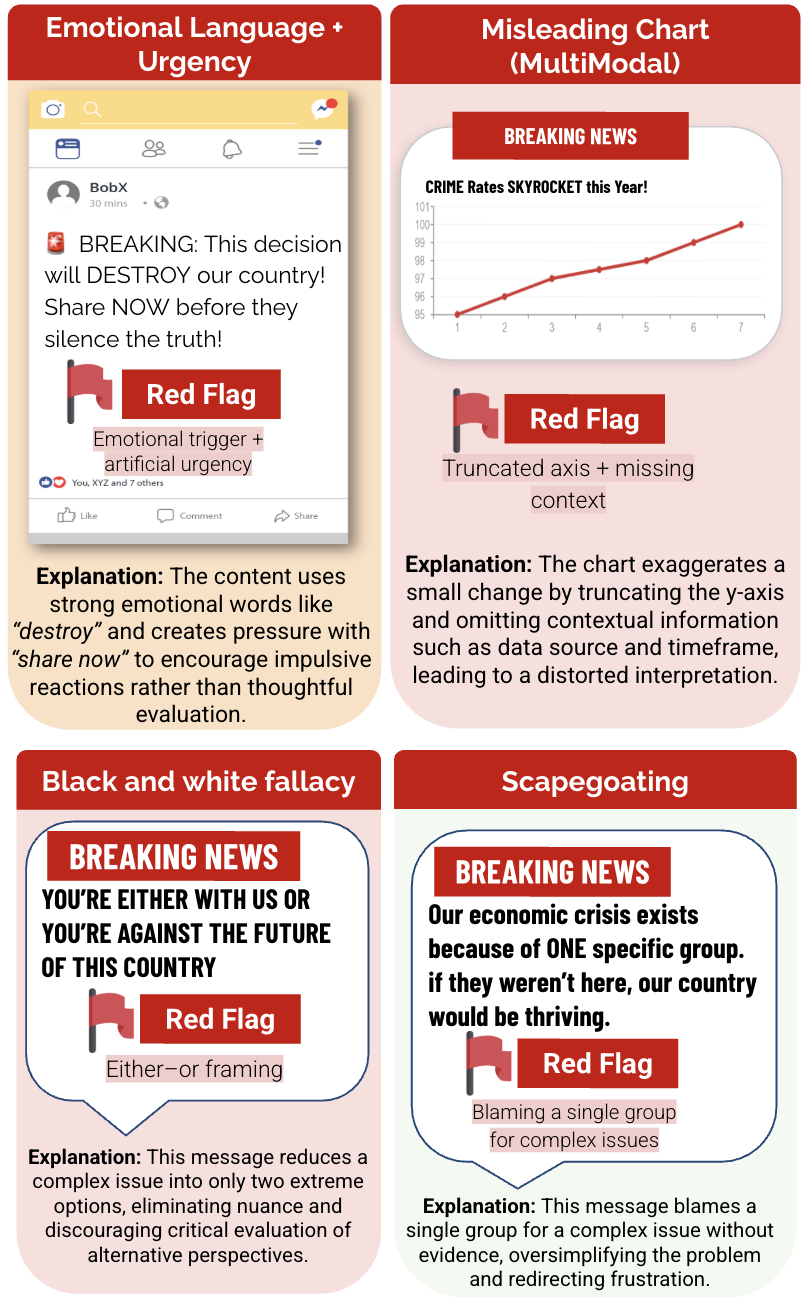}
    \vspace{-0.2cm}
    \caption{Examples of fictional social media posts, demonstrating different manipulation techniques.}    
    \label{fig:examples}
    \vspace{-0.4cm}
\end{figure}

    These limitations motivate complementary, user-centered approaches that improve \emph{critical evaluation} rather than relying on perfect detection coverage. Digital media literacy interventions can improve users’ ability to distinguish mainstream from false content, with effects that persist beyond immediate exposure \citep{guess-etal-2020-digitalmedia}. \emph{Prebunking} (psychological inoculation) extends media literacy by targeting manipulation \emph{techniques} (e.g., emotional (loaded) language, black white fallacy~\cite{da-san-martino-etal-2019-fine}, as shown in Figure \ref{fig:examples}).
Scalable inoculation interventions have been shown to increase resistance to misinformation tactics on social media \citep{roozenbeek-etal-2022-inoculation-social-media}. Meta-analyses further support improvements in credibility judgment across studies and settings \citep{lu-etal-2023-inoculation-meta,traberg-etal-2022-inoculation}.

Yet most media literacy and inoculation evaluations remain short, one-off web experiments, with limited attention to \textit{(i)} sustained engagement in everyday contexts, \textit{(ii)} multilinguality, and \textit{(iii)} durability and behavioral outcomes beyond immediate post-intervention assessments~\citep{lu-etal-2023-inoculation-meta,traberg-etal-2022-inoculation}. \textsc{CritiSense} is motivated by these gaps. It treats automatic detection as a safety net, but prioritizes building \emph{transferable} user competence through an iterative, mobile-first learning experience, along with an assessment method designed to measure not only immediate gains but also real-world impact.

Our contributions are as follows:
\begin{itemize}[noitemsep,topsep=0pt,leftmargin=*,labelsep=.5em]
    \item We introduce \textsc{CritiSense}, a \textbf{mobile-first micro-lessons} app that trains users to recognize misinformation and manipulation tactics in realistic, everyday scenarios.
    \item To our knowledge, this is the \textbf{\textit{first} multilingual} app of its kind, launched in Arabic and English and extended to Bangla, French, Hindi, Italian, Filipino, Nepali, and Urdu, offering full lesson content, quizzes, and feedback.
    \item We describe the complete user workflow, from learning and quizzes to simulated practice.
    \item We manually develop the learning content, covering core digital literacy concepts, propaganda techniques, and fake-news patterns.
    \item We integrate factuality, propaganda, and hate detection functionalities to provide automatic signals for textual and visual manipulation, supporting in-app feedback and learning.
    \item We report a formative usability evaluation measuring \textit{(i)} ease of use, \textit{(ii)} visual design, and \textit{(iii)} perceived content impact.
\end{itemize}

\noindent \textbf{Does CritiSense provide a usable and satisfying  experience?} A formative usability evaluation with 93 first-time users demonstrates strong overall usability (mean construct scores of 3.99–4.20 on a 5-point Likert scale), with 90.1\% of users agreeing that the app is easy to use and 83.9\% reporting satisfaction with their experience.

\section{\textsc{CritiSense} App}
\label{sec_critiSense_app}

\subsection{App Design}
\textsc{CritiSense} is a mobile-first media-literacy app designed to help users build and strengthen skills for evaluating online information. The design of the app is grounded in \textit{active learning}. Users read short lessons covering a wide variety of topics related to critical digital literacy, fake news and propaganda, answer short quizzes, receive immediate feedback, and practice applying the same reasoning to new examples. The app is also informed by \textit{cognitive inoculation}, training recognition of common manipulation techniques (e.g., loaded language, name calling) to increase users' resilience before they encounter them in the wild \cite{roozenbeek2020crosscultural}.

In Figure~\ref{fig:system_overview}, we provide an overview of \textsc{CritiSense}, organized around four core app components, and illustrate the end-to-end user journey from onboarding to practice. The app features a streamlined onboarding process: new users register and receive introductory guidance, while returning users authenticate directly. All pathways then converge on a dashboard/progress view, which branches into three functional modules: learning and knowledge assessment, interactive content verification, and simulated practice.

\noindent
\textbf{Dashboard/Progress.} This module displays the user profile, including the number of completed lessons, a score summary, and a leaderboard of top users with their ratings.

\noindent
\textbf{Learning/Play.} This module consists of lessons divided into chapters, organized by topic or manipulation technique. For example, the \textit{Critical Digital Literacy} lesson is divided into: \textit{(i)} the digital space and us, \textit{(ii)} developing critical digital literacy, and \textit{(iii)} critical digital literacy skills. Each chapter presents key definitions and concepts, followed by quizzes for reinforcement. All contents of the app are developed manually. 

\noindent
\textbf{Learning/Test Knowledge.} This module provides a flexible, non-linear self-assessment experience, allowing users to attempt targeted question banks and localized quizzes to gauge their understanding and earn rewards. The quizzes use multiple-choice questions to assess both conceptual knowledge and applied reasoning beyond individual lessons.

\noindent
\textbf{Detect.} This module serves as a practical verification tool where users can submit multimodal content through text input, image uploads, or direct URLs for real-time analysis. %
In addition to providing automatic predictions, it generates explanations that highlight linguistic cues, emotional framing, persuasive patterns, and other indicators of misleading information, helping users better understand why a piece of content may be unreliable.

\begin{figure}[]
    \centering    
    \includegraphics[width=0.98\columnwidth]{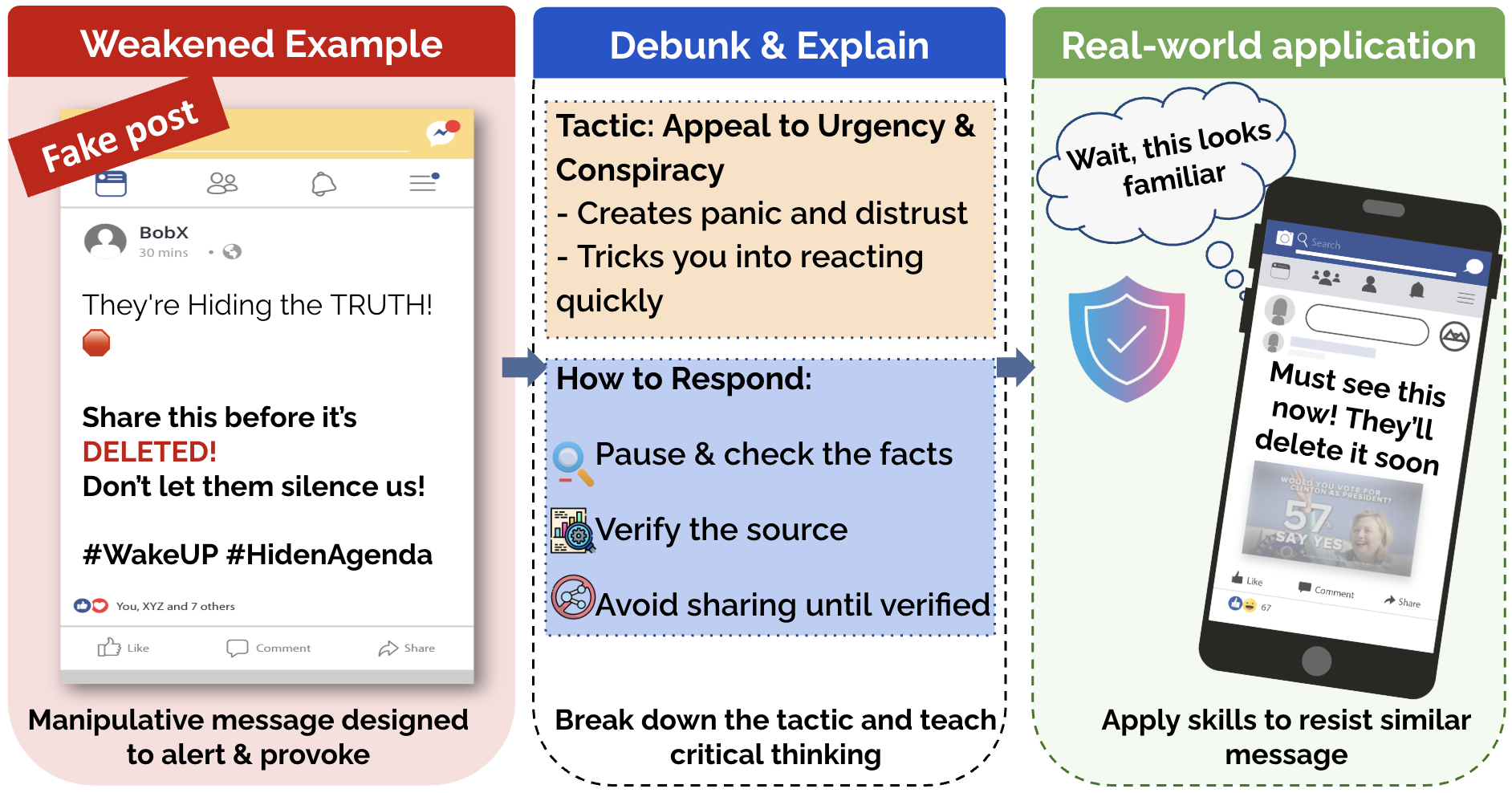}
    \vspace{-0.2cm}
    \caption{A pictorial example illustrating inoculation theory: expose a weakened misleading message, explain the tactic, then test transfer to a novel post.}    
    \label{fig:inoculation}
    \vspace{-0.3cm}
\end{figure}

\noindent
\textbf{Practice.} This module operationalizes inoculation theory~\cite{banas-rains-2010-inoculation-meta} through an interactive simulation titled \textit{``The Fake News Factory''} (Figure~\ref{fig:inoculation}). In this environment, users engage with a conversational system to practice identifying and understanding disinformation tactics through guided interaction with simplified, fictional examples and immediate feedback. By exposing users to weakened forms of misleading arguments paired with explanations and refutations, the module aims to build cognitive resistance to future persuasion attempts and strengthen users’ ability to recognize manipulation in the wild.
To mitigate dual-use risks, the Practice module is implemented as a fully scripted, branching dialogue rather than open-ended generation. All prompts, options, and feedback are pre-authored and curated for educational purposes. Users select from predefined choices illustrating manipulation tactics (e.g., emotional language, vague evidence, manufactured urgency), each followed by an explanation of \emph{why} the tactic is persuasive. The system does not generate novel disinformation or target real entities, ensuring that users learn to recognize manipulation patterns without producing deployable persuasive content.

\paragraph{Design goals.}
\textsc{CritiSense} is guided by several design goals. 
\textbf{First}, it adopts a \emph{technique-first} approach, training users to recognize recurring propaganda and misinformation strategies rather than focusing only on individual claims. This helps users apply what they learn across different topics and contexts.
\textit{Second}, it emphasizes practice in realistic formats: examples mirror the kinds of posts and media users encounter in everyday online settings. \textit{Third}, it provides immediate feedback with explanations, going beyond binary correctness to highlight the reasoning behind each decision. \textit{Fourth}, it is designed to be accessible and scalable through short microlearning lessons that require minimal setup and support learning beyond classroom contexts. 
\textit{Finally}, it is multilingual, expanding from its initial launch to nine supported languages: Arabic, English, Bangla, French, Hindi, Italian, Filipino, Nepali, and Urdu.

\subsection{Functionalities}
\textsc{CritiSense} provides: \textit{(i)} interactive micro-lessons with short quizzes, \textit{(ii)} technique-first prebunking modules that teach common manipulation tactics through examples, \textit{(iii)} immediate, explanation-rich feedback, and \textit{(iv)} prompts for repeatable verification habits (e.g., check sources, separate claims from evidence, and cross-check across outlets). The app also offers progress tracking and topic-level assessments for continued reinforcement. Finally, an AI-assisted analysis tool helps users examine external text and images for potential manipulative cues. In Table~\ref{tab:language_coverage}, we summarize the current functionality coverage across languages.

\begin{table}[h]
\centering
\setlength{\tabcolsep}{2pt}
\scalebox{0.70}{
\begin{tabular}{lccccccccc}
\toprule
\textbf{Module} & \textbf{EN} & \textbf{AR} & \textbf{BN} & \textbf{FR} & \textbf{HI} & \textbf{IT} & \textbf{TL} & \textbf{NE} & \textbf{UR} \\
\midrule
Lessons        & \avail & \avail & \avail & \avail & \avail & \avail & \avail & \avail & \avail \\
Quizzes        & \avail & \avail & \avail & \avail & \avail & \avail & \avail & \avail & \avail \\
Detect (text)  & \avail & \avail & \inprog & \inprog & \inprog & \inprog & \inprog & \inprog & \inprog \\
Detect (image) & \avail & \avail & \inprog & \inprog & \inprog & \inprog & \inprog & \inprog & \inprog \\
Practice       & \avail & \avail & \avail & \avail & \avail & \avail & \avail & \avail & \avail \\
\bottomrule
\end{tabular}
}
\vspace{-0.2cm}
\caption{Language coverage and status of the modules. EN=English, AR=Arabic, BN=Bangla, FR=French, HI=Hindi, IT=Italian, TL=Filipino (Tagalog), NE=Nepali, UR=Urdu. \availlegend~= available; \inproglegend~= in progress at the time of writing this paper.}
\label{tab:language_coverage}
\vspace{-0.4cm}
\end{table}

\subsection{Models}

\textsc{CritiSense} currently supports \textit{(i)} text-based fact-checking and propaganda detection, and \textit{(ii)} image-based propaganda detection for Arabic and hateful meme detection for English. These models are deployed through an in-house API platform.\footnote{\url{https://apihub.tanbih.org/docs}} 

At present, automated detection is available for English and Arabic, while instructional components, including lessons, quizzes, and the Practice module, are accessible across all nine supported languages (Table~\ref{tab:language_coverage}). Extending detection to additional languages requires curated factuality and propaganda resources, which remain uneven across languages. We aim to add these functionalities for other languages as part of our ongoing work.

\subsubsection{Datasets}
\noindent
\textbf{Factuality (text).} We train the factuality model on a curated collection of publicly available datasets in Arabic and English. For Arabic, we include AraFacts2~\cite{ali-etal-2021-arafacts}, ANS-Claim~\cite{khouja-2020-stance}, CT22Claim~\cite{nakov-etal-2022-checkthat-task1}, NewsCredibilityDataset~\cite{samdani2023arabic}, and COVID19Factuality~\cite{alam-etal-2021-fighting-covid}. For English, we include PolitiFact~\cite{clef-checkthat:2023:task3}, CT22T3\_Factuality~\cite{alam-etal-2021-fighting-covid}, CheckThat!-style misinformation datasets~\cite{clef-checkthat:2022:task3,shahi2021overview}, and AVeriTeC~\cite{schlichtkrull2023averitec}.

\noindent
\textbf{Propaganda (text).} For text-based propaganda detection, we use PropXplain~\cite{hasanain-etal-2025-propxplain}, which contains labeled instances in both Arabic and English.

\noindent
\textbf{Propaganda (image).} For image-based Arabic propaganda detection, we use ArMeme~\cite{alam2024armeme} (propagandistic vs.\ non-propagandistic), a meme-centric dataset.

\noindent
\textbf{Hateful memes (image).} For meme English hate detection, we use the Facebook Hateful Memes dataset~\cite{kiela2020hateful} (hate vs.\ non-hate).

\noindent
Table~\ref{tab:datasets} reports the data used for each task, grouped by modality and language, along with the train/ dev/ test split sizes. Across tasks, development splits are used for model selection and tuning, while test splits are held out for final evaluation.

\begin{table}[]
\centering
\setlength{\tabcolsep}{2pt} 
\scalebox{0.75}{
\begin{tabular}{@{}llrrr@{}}
\toprule
\multicolumn{1}{c}{\textbf{Task/Dataset}} & \multicolumn{1}{c}{\textbf{Modality}} & \multicolumn{1}{c}{\textbf{Train}} & \multicolumn{1}{c}{\textbf{Dev}} & \multicolumn{1}{c}{\textbf{Test}} \\ \midrule
Factuality (En) & Text & 467,830 & 67,389 & 134,594 \\
Factuality (Ar) & Text & 28,221 & 4,507 & 7,847 \\
Propaganda (En) & Text & 4,472 & 621 & 922 \\
Propaganda (Ar) & Text & 18,453 & 1,318 & 1,326 \\
ArMeme (Ar) & Image & 3,604 & 522 & 1,021 \\
Hateful meme (En) & Image & 8,500 & 540 & 2,000 \\ \bottomrule
\end{tabular}
}
\vspace{-0.2cm}
\caption{Dataset statistics for each task and modality, showing the number of instances in the train/dev/test splits used in our experiments.}
\label{tab:datasets}
\vspace{-0.45cm}
\end{table}

\subsubsection{Model Training and Evaluation}

\paragraph{Model Training.} We prioritize models that are low-cost to deploy in resource-constrained settings, including CPU-based servers. Accordingly, we use lightweight, widely adopted backbones. BERT-base for English~\cite{devlin2019bert}, AraBERT for Arabic~\cite{antoun2020arabert}, and ViT-B/16 for image-based classification models~\cite{dosovitskiy2021vit}. These choices provide a strong accuracy \textit{vs.} efficiency trade-off and allow seamless integration into our in-house API platform. Note that all trained models are binary classifiers.

\paragraph{Results.} Table~\ref{tab:results} reports performance across tasks. Overall, the text models perform well on factuality and propaganda detection. Arabic factuality is highest (Micro-F1=0.868), suggesting that the training data is well represented. English factuality is lower (Micro-F1=0.726), which likely reflects broader topic diversity and more varied claim formulations. Propaganda detection performs similarly across languages, with Micro-F1 scores of 0.772 for English and 0.762 for Arabic. However, the task remains challenging as misleading information often relies on subtle rhetorical framing, implicit meanings, and contextual interpretation.

In contrast, image-based hateful/propagandistic meme detection remains more challenging. ViT-B/16 achieves moderate Macro-F1 on ArMeme (0.554) and lower performance on Hateful Memes (0.507). This gap is expected, as memes often rely on implicit cultural context, sarcasm, and fine-grained image--text interactions that lightweight vision-only models struggle to capture. While stronger multimodal models exist (e.g., \citealp{shahroor2026memelens}), we adopt lightweight models to meet deployment and CPU-serving constraints. 

\paragraph{Error analysis.} We conducted an error analysis of the multimodal model outputs and found that most errors occur in memes requiring multimodal grounding, especially those with implicit references or tightly coupled text-image semantics. To address these issues, we are exploring two directions: \textit{(i)} integrating stronger multilingual vision-language models for more accurate inference, and \textit{(ii)} expanding the training data with additional annotated meme datasets. We will retain lightweight models as a fallback to support efficient deployment in resource-constrained environments.

\begin{table}[]
\centering
\setlength{\tabcolsep}{2pt} 
\scalebox{0.75}{
\begin{tabular}{@{}lllr@{}}
\toprule
\textbf{Task/Dataset} & \textbf{Metric} & \textbf{Model} & \multicolumn{1}{l}{\textbf{Performance}} \\ \midrule
Factuality (En) & Mi-F1 & BERT-Base & 0.726 \\
Factuality (Ar) & Mi-F1 & AraBERT & 0.868 \\
Propaganda (En) & Mi-F1 & BERT-Base & 0.772 \\
Propaganda (Ar) & Mi-F1 & AraBERT & 0.762 \\
ArMeme (Ar) & Ma-F1 & ViT-B/16 & 0.554 \\
Hateful meme (En) & Ma-F1 & ViT-B/16 & 0.507 \\ \bottomrule
\end{tabular}
}
\vspace{-0.2cm}
\caption{Model performance on each task and dataset, reporting the evaluation metric (Micro-F1 for text; Macro-F1 for image) and the backbone used.}
\label{tab:results}
\vspace{-0.4cm}
\end{table}

\section{Evaluation of \textsc{CritiSense}}
\label{sec_experiments}
We evaluate \textsc{CritiSense} through a structured usability study covering both interface usability and perceived learning/behavioral outcomes. The questionnaire measures five usability constructs, along with single-item ratings for overall satisfaction, ease of completing the lesson, quiz flow, intention to continue using the app, and likelihood of recommending it. We also collect brief contextual information and open-ended feedback to identify concrete design priorities. The study was conducted for both Arabic and English languages.

\noindent
\textbf{Participants.}
We recruited 93 participants and administered the study through SurveyMonkey. Participants were primarily aged 18--24 (68.8\%). English was the most common language preference (62.4\%), followed by bilingual English/Arabic use (20.4\%) and Arabic (17.2\%). The educational background of the participants was predominantly bachelor’s degree level (80.6\%), and their fields of study were skewed toward STEM disciplines. Participation was voluntary, however, participants received a \$14 voucher as modest compensation.

\noindent
\textbf{Instrument.} 
The instrument combines quantitative ratings with qualitative feedback (Section~\ref{app:survey}). It includes 17 five-point Likert items (strongly disagree-strongly agree), a five-point overall satisfaction item, a seven-point ease-of-completion rating for the lesson/quiz flow, and a Net Promoter Score (0-10).\footnote{\textit{How likely are you to recommend [product/app] to a friend or colleague?}~\cite{reichheld2003nps}} We also measured participants’ intention to continue and collected two multi-select responses on \textit{(i)} activities completed and \textit{(ii)} area of expertise or educational background. In addition, we included three open-ended questions on \textit{(i)} points of confusion, \textit{(ii)} highest-priority fixes, and \textit{(iii)} most-liked features.

\noindent
\textbf{Usability constructs and scoring.}
We group the 17 Likert items into five theoretically motivated constructs (Appendix Table~\ref{tab:usability-constructs}): \textbf{Usability/UX} (ease of use and perceived usefulness), \textbf{Visual Design} (aesthetics and clarity of visual elements), \textbf{Navigation} (ease of moving through the interface and finding content), \textbf{Content Effectiveness} (quality of examples/quizzes and cross-language consistency), and \textbf{Behavioral Impact} (perceived changes in critical evaluation behaviors). For each respondent, we compute each construct score as the mean of its constituent items, supporting both item-level analysis and construct-level comparisons.

\noindent
\textbf{Analysis.}
We compute descriptive statistics, including the mean (\textbf{M}), standard deviation (\textbf{SD}), and median, for all individual items and aggregated construct scores. We assess internal consistency for each construct using Cronbach's alpha ($\alpha$. Finally, we examine relationships among constructs using Pearson correlations ($r$)

\section{Findings}
\label{sec_findings}

\begin{table}[t]
\centering
\setlength{\tabcolsep}{2pt} 
\scalebox{0.75}{
\begin{tabular}{lccc}
\toprule
\textbf{Construct} & \textbf{Mean (SD)} & \textbf{$\alpha$} & \textbf{\% Positive} \\
\midrule
Usability / UX          & 4.12 (0.85) & 0.653 & 77.4\% \\
Visual Design           & 4.16 (0.70) & 0.774 & 72.0\% \\
Navigation              & 4.20 (0.84) & 0.837 & 73.1\% \\
Content Effectiveness   & 4.09 (0.62) & 0.746 & 65.6\% \\
Behavioral Impact       & 3.99 (0.72) & 0.840 & 63.4\% \\
\midrule
\textbf{Overall (17 items)} & --- & \textbf{0.921} & --- \\
\bottomrule
\end{tabular}
}
\vspace{-0.2cm}
\caption{Construct-level summary statistics, including mean (\textbf{M}), standard deviation (\textbf{SD}), internal consistency ($\alpha$), and the percentage of positive responses. Overall ratings for each questionnaire item in the construct were out of 5.}
\label{tab:construct_summary}
\vspace{-0.3cm}
\end{table}

\paragraph{Usability and satisfaction.}
Overall usability and satisfaction was strong (see Figure \ref{fig:satisfaction} and \ref{fig:easy_to_use} in Appendix). All five constructs exceeded the agreement threshold (M$>$3.0) as reported in Table \ref{tab:construct_summary}, with Navigation rated highest (M=4.20, SD=0.84) followed by Visual Design (M=4.16, SD=0.70). At the item level, \emph{easy to use} received the highest rating (M=4.43, SD=0.88), with 90.1\% of respondents selecting agree or strongly agree. Self-reported satisfaction was similarly high: 83.9\% of users indicated they were satisfied or very satisfied (M=4.14, SD=0.87).

\noindent
\textbf{Learning design and engagement.}
Participants responded positively to the pedagogical structure, particularly the quiz-based reinforcement. The item \emph{Quizzes reinforce learning} scored M=4.30 (SD=0.79), and 90.3\% of participants completed at least one quiz during their session. At the construct level, \emph{content effectiveness} was high (M=4.09, SD=0.62), suggesting that lessons communicated the targeted critical-thinking concepts. 

\noindent
\textbf{Behavioral impact.}
The \textit{behavioral impact construct} scored M=3.99 (SD=0.72), with 63.4\% of users providing positive ratings. Notably, Arabic-language users reported the highest behavioral impact (M=4.36), which is consistent with the hypothesis that the app might be valuable in contexts where localized and language specific critical-thinking resources are less available. 

\noindent
\textbf{Instrument reliability.}
It was high overall. Cronbach's $\alpha$ ranged from 0.653 for Usability/UX (2 items) to 0.840 for behavioral impact (4 items), with an overall scale reliability of $\alpha$=0.921. The lower value for the two-item construct is expected since $\alpha$ is sensitive to the number of items, while the full scale shows excellent reliability.

\noindent
\textbf{Qualitative feedback and improvement targets.}
Open-ended questions had high response rates (86--100\%), providing actionable feedback. Participants frequently highlighted \textit{ease of use} and the quiz-based learning flow. Responses also identified some issues that can help us to improve the app.

\noindent
\textbf{Interpretation and implications.}
Overall, results indicate that \textsc{CritiSense} provides a usable and engaging first-release learning experience, with clear priorities for iteration. Navigation was rated highest (M=4.20), suggesting that the chapter$\rightarrow$lesson$\rightarrow$quiz flow is easy to follow. Quiz-based reinforcement was also well received (M=4.30), and 90.3\% of participants completed at least one quiz.

Correlations are consistent with a ``UX$\rightarrow$content$\rightarrow$impact'' pattern. UX ratings correlate with content effectiveness ($r=0.58$), and content effectiveness correlates with behavioral impact ($r=0.63$). While not causal, these associations suggest that improving UX may yield downstream gains in perceived learning outcomes. Finally, NPS reflects an early-release profile (NPS$=-8.6$): 31.2\% of users are promoters (mode$=$10/10), %
with detractors accounting for 39.8\% of responses.

\paragraph{Pre/post app-use learning gains.}
To complement the usability findings, we conducted a preliminary pre/post app-use study through Prolific using misinformation and propaganda recognition quizzes. Among $n{=}52$ participants, accuracy in identifying misleading information increased from 70.2\% to 77.4\%, an absolute gain of $+7.2$ points and a relative improvement of $+10.3\%$.
The largest gains were observed for propaganda technique identification. Each participant received \pounds8 as compensation. A full longitudinal study with a control group, %
and behavioral measures is ongoing and will be reported in future work.

\section{Related Work}
\label{sec:related_work}

\noindent
\textbf{Detection-based and platform interventions.}
There has been substantial amount of work addressing misinformation through automatic detection, fact-checking pipelines, and platform UI interventions (e.g., labels and warnings) \cite{hasanain-etal-2024-large,hasanain-etal-2024-gpt,alam-etal-2022-survey,alam-etal-2021-fighting-covid,zhou2020survey,shaar-etal-2022-assisting,abouzied2025combating}. While these systems provide important first-line safeguards, their effectiveness can degrade under temporal shift~\cite{stepanova-ross-2023-temporal}, vary across languages in cross-lingual transfer~\cite{ozcelik-etal-2023-cross}, and yield limited average changes in real-world beliefs or consumption~\cite{aslett2022_labels}. Moreover, selective labeling can increase perceived accuracy of untagged content (the ``implied truth'' effect)~\cite{pennycook-etal-2020-implied-truth}.

\noindent
\paragraph{Prebunking and inoculation.}
Psychological inoculation builds resistance by teaching manipulation techniques rather than correcting individual claims. Games such as \emph{Bad News} and \emph{Harmony Square} improve recognition of propaganda tactics and reduce perceived credibility of misleading content~\cite{roozenbeek2019_fake_news_game,roozenbeek2020_harmony_square}, with similar effects reported in topic-focused and multilingual variants~\cite{basol2021_covid_prebunk}. Scalable prebunking videos can also improve technique recognition on social media~\cite{roozenbeek-etal-2022-inoculation-social-media}. However, many interventions are delivered as one-off web experiences and are evaluated primarily with immediate post-intervention assessments.

\noindent
\paragraph{Media literacy tools.}
Mobile-first media literacy tools remain comparatively less common. \emph{Cranky Uncle} uses humor and fallacy training to build resilience against climate misinformation~\cite{cook2023_cranky_uncle}, while feed simulations such as \emph{Fakey} provide ecologically realistic practice and report improved source discernment in longitudinal deployments~\cite{micallef2021_fakey}. Overall, prior tools tend to emphasize either short-session tactic inoculation or practice-based news literacy, but rarely combine both within a multilingual, mobile-first experience designed for sustained everyday use beyond classroom settings.

\noindent
\paragraph{\textsc{CritiSense}.}
\textsc{CritiSense} bridges these strands by combining tactic-level prebunking with practical verification skills (e.g., distinguishing opinion from evidence, fallacy spotting, and verification habits) through short, interactive exercises with explanation-rich feedback. Unlike primarily browser-based prebunking tools, it is mobile-first, multilingual, and designed for repeated microlearning. 

More concretely, CritiSense differs from prior prebunking and media-literacy tools along three axes. \textit{First}, browser-based inoculation games such as \emph{Bad News}~\cite{roozenbeek2019fake} and \emph{Harmony Square}~\cite{roozenbeek2020_harmony_square} are typically delivered as one-off sessions, whereas CritiSense supports continuous, repeated engagement through microlearning. \textit{Second}, tools such as \emph{Cranky Uncle}~\cite{cook2023_cranky_uncle} and \emph{Fakey}~\cite{micallef2021_fakey} are often limited to specific domains or primarily English-language settings, while \textsc{CritiSense} provides consistent lesson, quiz, and Practice functionality across nine languages. \textit{Third}, whereas prior systems commonly emphasize either tactic-level inoculation or practice-based literacy, \textsc{CritiSense} integrates both with explanation-rich feedback and an AI-assisted detection module within a single deployed platform.

\section{Conclusions}
\label{sec_conclusions}
We presented \textsc{CritiSense}, a mobile-first media-literacy app that delivers technique-focused prebunking and practical verification skills through short, interactive exercises with explanation-rich feedback. \textsc{CritiSense} is, to our knowledge, the first effort of this kind released in multiple languages, starting with Arabic and English and extending to additional seven languages. A formative usability study with 93 users shows strong user experience, including high usability (Navigation M=4.20; ease-of-use M=4.43), high satisfaction (83.9\% positive), excellent internal consistency of the evaluation instrument (Cronbach's $\alpha=0.92$) $+10.3\%$ relative) gain in quiz accuracy after app use. Qualitative feedback further validates the chapter$\rightarrow$quiz learning loop and highlights functionality improvements. Overall, these findings support \textsc{CritiSense} as a scalable multilingual prebunking platform and motivate future work on long-term learning, behavioral impact, and multimodal analysis.

\paragraph{Limitations.}
\textsc{CritiSense} is designed to strengthen awareness and skills for recognizing fake news, mis/disinformation, and propaganda. Our current evaluation is a pilot mixed-method study focused on usability, engagement, and short-term learning signals; it does not yet measure long-term retention or real-world behavioral change (e.g., sharing on users’ own platforms). The app presently covers a curated set of techniques and item types, and the content requires ongoing updates and localization. 

To support scalability, the platform adopts a modular design: lessons, quizzes, and Practice scenarios are stored as structured content, allowing rapid updates without requiring app-store re-releases. We are also piloting an LLM-assisted authoring workflow, where generated lesson drafts are curated and validated by expert reviewers, enabling faster adaptation to emerging narratives while maintaining content quality.

While the design aims to generalize across topics, we do not claim coverage across all domains or user groups, and larger longitudinal and cross-cultural studies are needed. Finally, \textsc{CritiSense} is intended to complement platform detection and fact-checking systems not to replace them.

\paragraph{Ethics and broader impact.}
\textsc{CritiSense} operates in a sensitive domain where design choices can inadvertently amplify harmful narratives. To mitigate this, the app teaches manipulation patterns (e.g., emotional framing, scapegoating) using a technique-centered approach and avoids presenting harmful misinformation. Feedback emphasizes actionable verification steps (e.g., source checks, evidence tracing) rather than restating false claims. We also acknowledge dual-use risks: explanations of propaganda strategies could be misused to craft persuasive misinformation. \textsc{CritiSense} reduces this risk by prioritizing recognition and critical questioning, not step-by-step guidance for manipulation, and by curating examples for educational intent. Regarding privacy, \textsc{CritiSense} does not collect sensitive user information. Overall, we expect positive impact. Improved media literacy can support informed participation and reduce susceptibility to manipulation, particularly in multilingual settings. We plan longitudinal studies to assess durability and monitor potential unintended effects (e.g., overconfidence or blanket skepticism).

\section*{Acknowledgments}
The work was supported by NPRP grant 14C-0916-210015 from the Qatar National Research Fund, part of the Qatar Research Development and Innovation Council (QRDI). The findings reported herein are solely the responsibility of the authors.

\bibliography{bibliography/bibliography}

\appendix

\label{sec:appendix}
\balance
\section{Usability constructs}

Table~\ref{tab:usability-constructs} summarizes the five usability constructs used in our evaluation and the number of Likert items mapped to each construct. These constructs span usability/UX, visual design, navigation, content effectiveness, and behavioral impact, and are used to compute construct-level scores by averaging their constituent items.

\begin{table}[h]
\centering
\setlength{\tabcolsep}{2pt} 
\scalebox{0.72}{
\begin{tabular}{p{1.8cm}p{1.2cm}p{6.7cm}}
\hline
\textbf{Construct} & \textbf{\#} & \textbf{Description} \\
\hline
\textbf{Usability / UX} & 2 &
General usability: capabilities meet needs; ease of use \\
\textbf{Visual Design} & 4 &
Aesthetics and clarity: design appeal; readability; icon clarity; animation clarity \\
\textbf{Navigation} & 3 &
Ease of navigation; finding lessons; speed and responsiveness \\
\textbf{Content Effectiveness} & 4 &
Misinformation identification; example relevance; quiz reinforcement; EN/AR consistency \\
\textbf{Behavioral Impact} & 4 &
Confidence evaluating information; questioning reliability; noticing manipulation tactics; helping others \\
\hline
\end{tabular}
}
\vspace{-0.2cm}
\caption{Usability constructs and item counts used in the evaluation.}
\label{tab:usability-constructs}
\vspace{-0.3cm}
\end{table}

\section{Usability Study Questionnaire}
\label{app:survey}

The following questionnaire was administered to \(N = 93\) participants via SurveyMonkey after interacting with the CritiSense application.
All Likert items used a 5-point scale: \textit{Strongly Disagree} (1) -- \textit{Strongly Agree} (5), unless stated otherwise.

\subsection*{Section 1: Consent \& App Usage}

\begin{enumerate}[label=\textbf{Q\arabic*.}]

  \item Which device did you use for CritiSense?\\
        \textit{[Android phone / iPhone / Tablet / Computer]}

  \item Which language did you use in the app?\\
        \textit{[English / Arabic / Both English \& Arabic]}

  \item Is this your first time using CritiSense?\\
        \textit{[Yes / No]}

  \item Which activities did you complete? \textit{(Select all that apply)}\\
        \textit{[Browsed the home screen / Completed a quiz / Switched app language / Explored multiple sections / Other]}

  \item What feature do you see in the app after reading a chapter?\\
        \textit{[Multiple choice: questions related to the chapter / another lesson / another chapter]}

\end{enumerate}

\subsection*{Section 2: Usability \& UX}

\begin{enumerate}[resume, label=\textbf{Q\arabic*.}]

  \item CritiSense's technical capabilities meet my needs.
  \item CritiSense is easy to use.
  \item Overall, how easy or difficult was it to complete a lesson and quiz?\\
        \textit{[1 = Extremely difficult \quad 7 = Extremely easy]}

\end{enumerate}

\subsection*{Section 3: Visual Design}

\begin{enumerate}[resume, label=\textbf{Q\arabic*.}]

  \item The visual design of the app is appealing.
  \item The text is easy to read (size, contrast, spacing).
  \item Icons and interface elements are clear and understandable.
  \item Animations and transitions add clarity to the experience.

\end{enumerate}

\subsection*{Section 4: Navigation}

\begin{enumerate}[resume, label=\textbf{Q\arabic*.}]

  \item It is easy to navigate through the app.
  \item I can find lessons and quizzes easily.
  \item The app feels fast and responsive.

\end{enumerate}

\subsection*{Section 5: Content Effectiveness}

\begin{enumerate}[resume, label=\textbf{Q\arabic*.}]

  \item The lessons helped me better identify misinformation and manipulation techniques.
  \item The examples used in the lessons feel relevant to real situations.
  \item The quizzes reinforce what I learned.
  \item The English and Arabic content feel consistent in quality and coverage.

\end{enumerate}

\subsection*{Section 6: Behavioral Impact}

\begin{enumerate}[resume, label=\textbf{Q\arabic*.}]

  \item After using CritiSense, I feel more confident evaluating information I see online.
  \item I am more likely to question the reliability of social media posts after using CritiSense.
  \item CritiSense helped me notice common tricks used to manipulate people online.
  \item After using CritiSense, I am more likely to help friends or family spot misleading or manipulative online content.

\end{enumerate}

\subsection*{Section 7: Overall Evaluation}

\begin{enumerate}[resume, label=\textbf{Q\arabic*.}]

  \item Overall, how satisfied are you with CritiSense?\\
        \textit{[1 = Very Dissatisfied \quad 5 = Very Satisfied]}

  \item I would like to continue using CritiSense in the future.

  \item How likely are you to recommend CritiSense to a friend or colleague?\\
        \textit{[0 = Not at all likely \quad 10 = Extremely likely]}

\end{enumerate}

\subsection*{Section 8: Open-Ended Feedback}

\begin{enumerate}[resume, label=\textbf{Q\arabic*.}]

  \item What part of the app (if any) felt confusing or difficult?
  \item If we fix one thing first, what should it be?
  \item What did you like most about using CritiSense?

\end{enumerate}

\subsection*{Section 9: Demographics}

\begin{enumerate}[resume, label=\textbf{Q\arabic*.}]

  \item What is your highest level of education?\\
        \textit{[High school or equivalent / Some college / Bachelor's (pursuing/completed) /
        Master's (pursuing/completed) / PhD or Doctorate / Other]}

  \item What is your field of study or expertise? \textit{(Select all that apply)}\\
        \textit{[Education / Media \& Communication / Technology \& IT / Business \& Management /
        Engineering / Health \& Medicine / Social Sciences / Prefer not to say / Other]}

  \item What is your age group?\\
        \textit{[Under 18 / 18--24 / 25--34 / 35--44 / 45--54 / 55+]}

\end{enumerate}

\section{Usability and Satisfaction}
\begin{figure}[!tbh]
    \centering    
    \includegraphics[width=0.85\columnwidth]{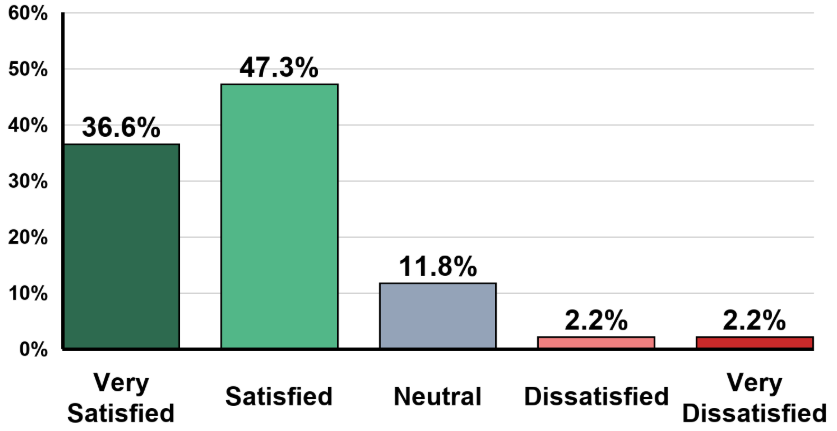}
    \vspace{-0.2cm}
    \caption{User satisfaction survey results (N=93). The distribution indicates high platform acceptance, with 83.9\% of participants reporting positive sentiment (36.6\% Very Satisfied and 47.3\% Satisfied). Neutral responses accounted for 11.8\%, while combined negative sentiment (Dissatisfied and Very Dissatisfied) remained minimal at 4.4\%.}    
    \label{fig:satisfaction}
\end{figure}

\begin{figure}[!tbh]
    \centering    
    \includegraphics[width=0.9\columnwidth]{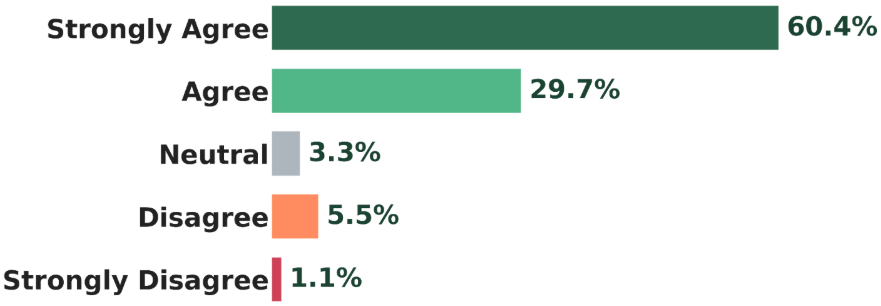}
    \caption{User perception of system usability ($N=93$). A significant majority (90.1\%) expressed agreement, with 60.4\% indicating they ``Strongly Agree.''}
    \label{fig:easy_to_use}
    \vspace{-0.3cm}
\end{figure}

Figures~\ref{fig:satisfaction} and~\ref{fig:easy_to_use} summarize participants' overall satisfaction with \textsc{CritiSense} and their perceived \textit{ease of use}. The satisfaction distribution (Figure~\ref{fig:satisfaction}) is strongly skewed toward positive responses: 36.6\% of participants reported being \emph{Very Satisfied} and 47.3\% \emph{Satisfied}, yielding a combined positive sentiment of 83.9\%. Neutral responses accounted for 11.8\%, while negative responses were marginal at 4.4\% combined (2.2\% \emph{Dissatisfied} and 2.2\% \emph{Very Dissatisfied}). This pattern indicates broad platform acceptance among first-time users across both Arabic and English participants.

Perceived \textit{ease of use} (Figure~\ref{fig:easy_to_use}) follows a similar but even stronger positive skew. A combined 90.1\% of participants agreed the app was \textit{easy to use}, with 60.4\% selecting \emph{Strongly Agree} and 29.7\% \emph{Agree}. Neutral responses were minimal at 3.3\%, and disagreement was limited to 5.5\% \emph{Disagree} and 1.1\% \emph{Strongly Disagree}. The clustering of responses at the \emph{Strongly Agree} end suggests that the chapter$\rightarrow$lesson$\rightarrow$quiz workflow and onboarding flow lower the entry barrier for new users. Taken together, these distributions corroborate the construct-level results in Table~\ref{tab:construct_summary}, where Usability/UX, Visual Design, and Navigation all exceeded a mean score of 4.1 out of 5.

\end{document}